\documentclass[letterpaper, 10 pt, conference]{ieeeconf}  

\usepackage{graphicx} 
\usepackage{amsmath} 

\usepackage{caption}
\usepackage{subcaption}

\IEEEoverridecommandlockouts                              
\title{\LARGE \bf
Learning an internal representation of the end-effector configuration space
}

\author{Alban Laflaqui\`ere$^{1}$, Alexander V. Terekhov$^{1}$, Bruno Gas$^{1}$ and J. Kevin O'Regan$^{2}$
\thanks{* This work was conducted within the French/Japan BINAAHR (BINaural Active Audition for Humanoid Robots) project under Contract n$^\circ$ \mbox{ANR-09-BLAN-0370-02} funded by the French National Research Agency.}
\thanks{$^{1}$ is with UPMC Univ Paris 06 and ISIR (CNRS UMR 7222), F-75005, Paris, FRANCE
              {\tt\small name@isir.upmc.fr}}%
\thanks{$^{2}$ is with Univ Paris 05 Descartes and LPP (CNRS UMR 8158), F-75006, Paris, FRANCE
              {\tt\small jkevin.oregan@gmail.com}}%
}

\begin{document}

\maketitle
\thispagestyle{empty}
\pagestyle{empty}

\begin{abstract}

Current machine learning techniques proposed to automatically discover a robot kinematics usually rely on \textit{a priori} information about the robot's structure, sensors properties or end-effector position. This paper proposes a method to estimate a certain aspect of the forward kinematics model with no such information. An internal representation of the end-effector configuration is generated from unstructured proprioceptive and exteroceptive data flow under very limited assumptions. A mapping from the proprioceptive space to this representational space can then be used to control the robot.
\end{abstract}

\section{INTRODUCTION}



One of the problems an autonomous robot must be able to solve is to retrieve basic information about its own topological structure relying on minimal \textit{a priori} information. The problem of learning the kinematics of a manipulator was approached by a number of machine learning techniques (see review in \cite{Sigaud2011}). It is usually assumed that the end-effector position is registered using an external measuring system and the problem consists of finding a mapping between the joints space and the end-effector position. Some attempts were made to learn so-called 'visual-motor' mappings between the joint angles and end-effector's image in the robot-mounted visual system \cite{Droniou2012}. These approaches rely a lot on \textit{a priori} assumptions about the properties of sensors and the robot structure. In alternative approaches no such assumptions are made, but they require the robot surface to be covered with an artificial skin \cite{Schatz2009,McGregor2011}.

Imagine a robotic arm with an unspecified number of links and a camera installed on one of them. The camera output is scrambled, so that the order of pixels is completely broken and some of them are missing. The robot measures its configuration with uncalibrated encoders in the joints, but as the geometrical properties and links topology are unknown, these measurements do not tell anything about the actual configuration of the robot in space. Such a robot does not have explicit measure of its end-effector position, but this information is implicitly available if the output of the camera is different for different positions / orientations. This implicit information can be sufficient to build an internal representation of the end-effector space, which has the same topology as the actual end-effector space but may have different metrics. Such a representation simply associates all configurations of the robot, for which the camera position and orientation are the same and thus introduces a new topology on the space of the encoder outputs. Building such a representation would be a trivial task if the camera position and orientation were available. In our case they are only accessible through the scrambled picture it provides. In this study we present an algorithm that builds an internal representation of the end-effector configuration space, from the unordered and scrambled information provided by the camera (or any other exteroceptor) installed on the end-effector.

The overall idea is illustrated in Fig. 1. The robot accesses its kinematic state through the proprioception (e.g. encoders in joints) but has no information about its geometry or topological structure. The state of the environment is reflected in the outputs of the exteroceptors (camera, microphone array, set of photodiodes, etc). When the environment does not change, the output of the exteroceptors is determined by the position of the end-effector. The aim of the algorithm is to build an internal representation of the end-effector space by associating all proprioceptive inputs that correspond to the same outputs of the exteroceptor. Note that the topology of the internal representation can be different from the one of the true end-effector space if the exteroceptors have the same outputs for different end-effector positions.

We present a neural network-based algorithm that maps a potentially high dimensional proprioception space into a low dimensional space of internal representation. The dimension of the latter is assumed known here and can be estimated using methods like in \cite{Philipona2003,Laflaquiere2012}. We test the quality of the mapping by making a simulated robot performing reaching movements using the Jacobian of the mapping defined by the neural network.

\begin{figure}[h]
\centering
\includegraphics[width=\linewidth]{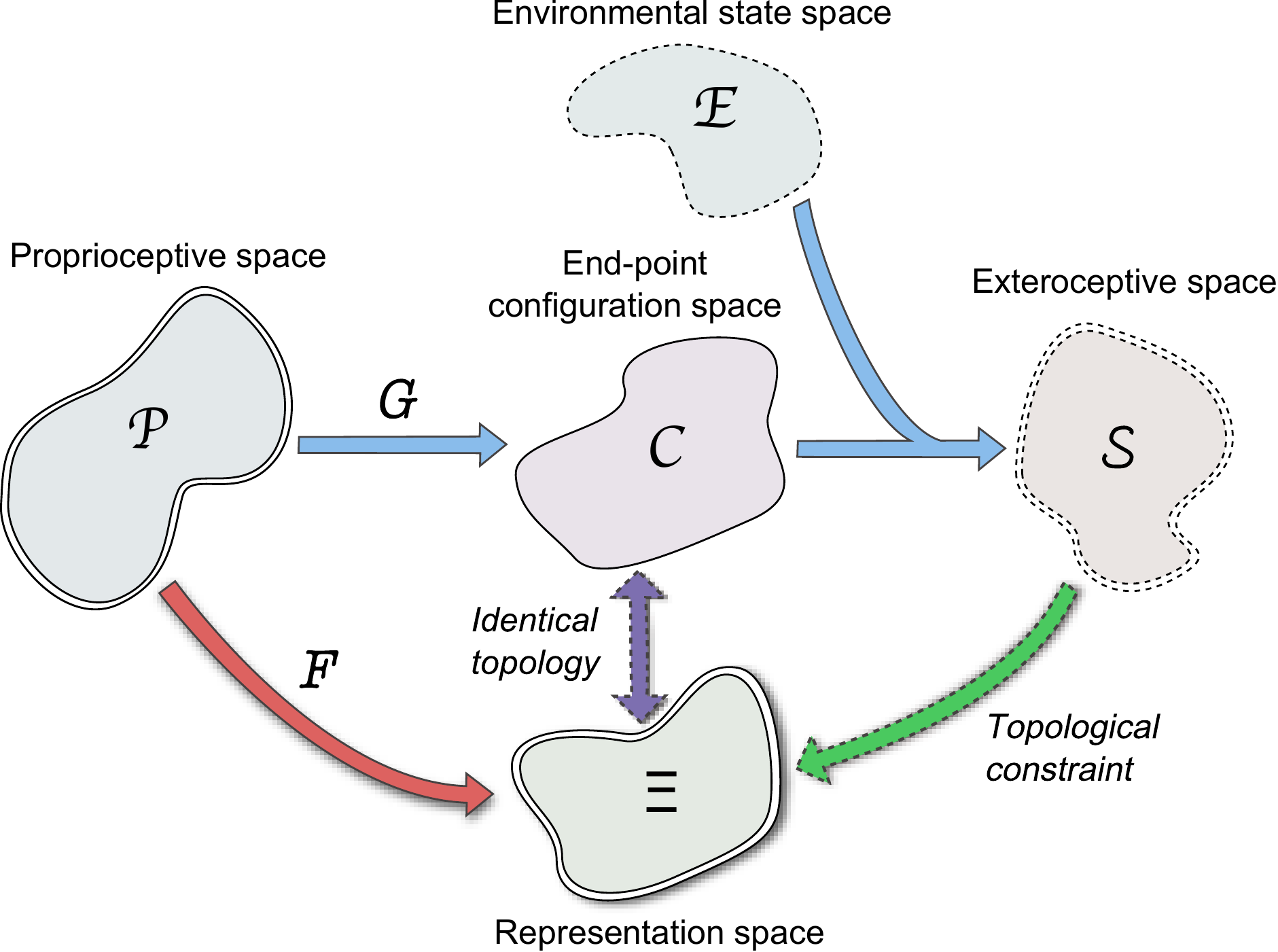}
\caption{Double-lined spaces are accessible to the robot. Dotted-lined spaces may change for different environments. Each proprioceptive state $p$ generates an end-effector configuration $c$ associated with a exteroceptive state $s$ that depends on the environmental configuration. The goal of the robot is to create an internal representation $\xi$ of the end-effector configuration using topological information of the exteroceptive space.}
\label{fig:spaces_and_mappings}
\end{figure}

\section{METHOD OVERVIEW}

The overall method section consists of three parts. The first part describes a learning algorithm for the neural network (NN) mapping the proprioception $\mathcal{P}$ to an internal representation $\Xi$, which is topologically equivalent to the space of end-effector configurations $\mathcal{C}$. The Jacobian of this mapping is computed in the second part. In the third part, the routine used for the reaching movement is described.

\subsection{Neural network and cost function}
\label{sec:Neural network and cost function}

The goal of the NN is to map elements $p$ of the proprioception space $\mathcal{P}$ into the elements $\xi$ of an \textit{a priori} unspecified internal representation space $\Xi$. The only constraint on this space is that it must be topologically identical to the space of end-effector configurations $\mathcal{C}$. As the end-effector configuration $c$ is unknown to the robot, the difference in the exteroceptors $s$ is used instead to generate $\Xi$. This approach implies that the robot sensory experience is not ambiguous, \textit{i.e.} each exteroceptive state $s$ is uniquely related to one configuration $c$ of the end-effector. Each configuration $c$ can however be related to multiple proprioceptive states $p$ if the robot is redundant.

\subsubsection{Neural network architecture}
We used a multilayer perceptron (MLP \cite{Rumelhart1986}) with one hidden layer and a sigmoid excitation function $\sigma$. Each weight $w_{l,k}$ is initially drawn from an uniform distribution in $[-1,1]$. Excitation $y_k$ of neuron $k$ is computed as follows:
\begin{equation}
  y_k = \sigma \Big( \sum_l w_{l,k} y_l \Big) = \sigma \big( e_k \big)
\end{equation}
with $y_l$ the excitation of neuron $l$ in previous layer and $\sigma(\cdot)$ the sigmoid function.

The number of input neurons $N^{in}$ is equal to the dimension of the proprioceptive space $\mathcal{P}$. The number of hidden neurons $N^h$ is arbitrarily set to $30$. It could naturally be modified according to the expected mapping complexity. The number of output neurons $N^{out}$ is equal to the number of independent variables necessary to describe the exteroceptive state $s$. In the following, $N^{out}$ is assumed known, but its value could be determined through an exteroceptive manifold analysis \cite{Philipona2003,Laflaquiere2012}. Note that if the robot's sensory capacity doesn't allow a complete characterization of the end-effector configuration $c$, the number of independent variables $N^{out}$ can be smaller than expected from an external point of view.

\subsubsection{Cost function minimization}
\label{sec:Cost function minimization}
As it is difficult to build a cost function that only captures the similarity of the topology, we required the NN to preserve in $\Xi$ the distances between the points in the exteroception space $\mathcal{S}$. The cost function is:
\begin{equation}
	\begin{array}{c}
	Q = \sum_{i,j} Q^{i,j}, \\
	Q^{i,j}  = h \Big( ||s^{i,j}|| \Big) \Big( ||\xi^{i,j}|| - ||s^{i,j}|| \Big)^2 + \gamma \Big( ||\xi^i||^2+||\xi^j||^2 \Big)\\
	\text{with:}\\
	||s^{i,j}|| = ||s^i -s^j|| \text{ and } ||\xi^{i,j}|| = ||\xi^i-\xi^j||
	\end{array}
\label{eq:cost}
\end{equation}
where $||.||$ denotes the euclidean norm, $\xi^i$ (resp. $\xi^j$) is the network output for the proprioceptive configuration $p^i$ (resp. $p^j$), $s^i$ (resp. $s^j$) is the exteroceptive state associated with $p^i$ (resp. $p^j$), $\gamma$ is a small centering factor added for the purpose of regularization, and $h(.)$ is a neighborhood function designed to give more weight to small exteroceptive distances. In the following, $h(.)$ is a simple step function:
\begin{equation}
	h(a) = \left\{
	\begin{array}{l l}
		1 & \text{ if } a \leq \mu\\
		0 & \text{ otherwize}
	\end{array}
	\right.
\end{equation}

Unlike traditional MLP cost functions, $Q^{i,j}$ is defined for pairs of inputs $\{p^i,p^j\}$ and outputs $\{\xi^i,\xi^j\}$. Only distances $||\xi^{i,j}||$ in the output space $\Xi$ are then constrained. Moreover, this constraint is defined through auxiliary distances $||s^{i,j}||$. The first term of the sum is minimized when outputs distances $||\xi^{i,j}||$ tend to be equal to exteroceptive distances $||s^{i,j}||$. The second term is a centering cost that is minimized when outputs $\xi$ tend towards $0$ (Note that $\gamma$ must be small enough not to disturb the first term minimization). Conservation of the exteroceptive space topology is achieved by respecting only small exteroceptive distances $||s^{i,j}||$ through the use of the neighborhood function $h(.)$.

We used RPROP algorithm to find a minimum of the cost function \cite{Riedmiller1993}. The algorithm updates the weights $w_{l,k}$ according to the sign of the local gradient $\frac{\partial Q^{i,j}}{\partial w_{l,k}}$ defined as follows:

\begin{itemize}
\item if neuron $k$ is in the output layer
\begin{align}
	&\frac{\partial Q^{i,j}}{\partial w_{l,k}} = h \big( ||s^{i,j}|| \big) \Big( \delta^i_k y^i_l + \delta^j_k y^j_l \Big) \\
	&\delta^i_k = 2\bigg( \gamma\big(y^i_k+y^j_k\big) + \big(y^i_k - y^j_k\big) \Big(1- \frac{||s^{i,j}||}{||\xi^{i,j}||}\Big) \bigg)\sigma'(e^i_k) \nonumber \\
	&\delta^j_k = 2\bigg( \gamma\big(y^i_k+y^j_k\big) - \big(y^i_k - y^j_k\big) \Big(1- \frac{||s^{i,j}||}{||\xi^{i,j}||}\Big)  \bigg)\sigma'(e^j_k) \nonumber
\end{align} 
\item if neuron $k$ is in the hidden layer
\begin{equation}
	\frac{\partial Q^{i,j}}{\partial w_{l,k}} = y^i_l \sigma'(e^i_k) \sum_n \delta^i_n w_{k,n} + y^j_l \sigma'(e^j_k) \sum_n \delta^j_n w_{k,n}
	\end{equation}
\end{itemize}
where $\sigma'(.)$ is the derivative of the excitation function (sigmoid).

\subsubsection{Exploration and learning}
The data necessary to perform the learning are generated through a random exploration of the robot's working space. First, $N$ proprioceptive states $p^i$ are randomly selected in the neighborhood of the reference configuration. The neighborhood was defined so that the deviation of each proprioceptive value $p^i_k$ did not exceed a predefined value $A$. The configurations falling out of the working space were discarded. Then the exteroceptive values  $s^i$ associated with those proprioceptive states are computed. The proprioceptive commands $p^i$ and exteroceptive states $s^i$ are normalized to avoid saturation of the sigmoid excitation function:
\begin{align}
	\tilde{p}^i &= 0.8 (p^i-\bar{p}) / \max_{i,k}(|p^i_k-\bar{p}_k|) \\
	\tilde{s}^i &= 0.4 (s^i-\bar{s}) / \max_{i,k}(|s^i_k-\bar{s}_k|)
\end{align}
Finally, all possible pairs $\{\tilde{p}^i,\tilde{p}^j\}$ with $j\in\{i+1,N\}$, and corresponding $\{\tilde{s}^i,\tilde{s}^j\}$, are created to form the learning base. The RPROP algorithm is applied for $T$ iterations or until the error $Q$ reaches a minimal value $Q_{min}$.

In most cases the cost function (\ref{eq:cost}) has numerous local minima, that makes its  minimization difficult. In order to assure better initial conditions we performed specific initialization of the network. We generated an $N^{out}$-dimensional unfolding of the exteroceptive set $\{\tilde{s}^i\}$ using Isomap \cite{Tenenbaum2000}. A usual quadratic error minimization is then performed using the resulting projection as the desired values $\xi$ of the NN \cite{Rumelhart1986}.

This preliminary learning is supposed to bring the network in a favourable basin of attraction before the less-constrained cost minimization described in \ref{sec:Cost function minimization} is applied. Note that the NN initialization is performed for a fixed environment. The latter can however be different and even change during the $Q$ minimization as long as distances $||s^i-s^j||$ are computed for a single environment (see (\ref{eq:cost})). In the following, three random environments are considered, leading to three successive explorations during the minimization of $Q$.

\begin{figure*}[!t]
\centering
     \begin{subfigure}[b]{0.4\linewidth}
	\centering
	\includegraphics[width=\textwidth]{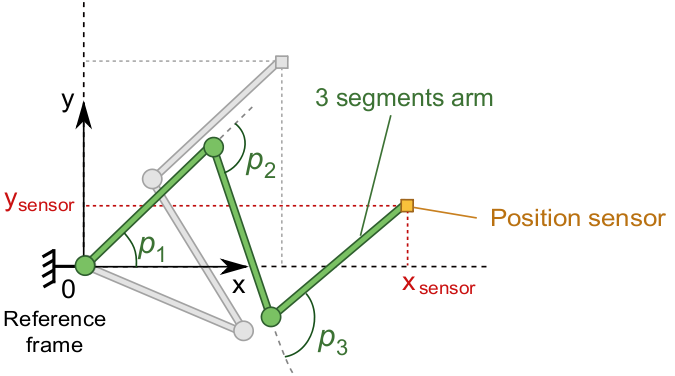}
	\caption{Scenario 1: position sensor.}
	\label{fig:arm}
	\end{subfigure}
	\quad
     \begin{subfigure}[b]{0.5\linewidth}
	\centering
	\includegraphics[width=\textwidth]{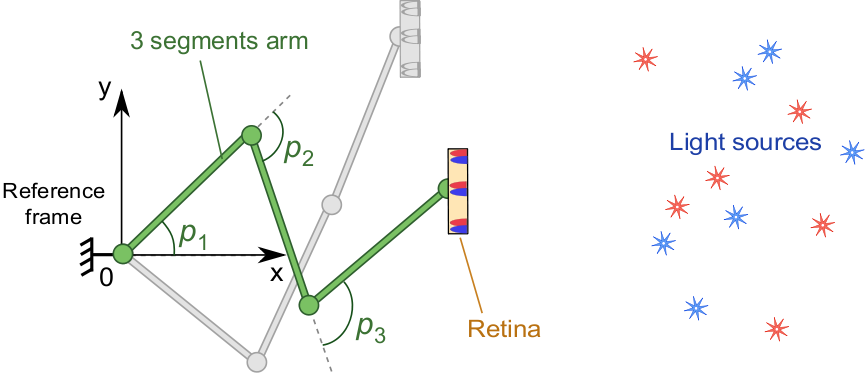}
	\caption{Scenario 2: retina.}
	\label{fig:arm2}
	\end{subfigure} \\
	\vspace{0.2cm}
     \begin{subfigure}[b]{0.423\linewidth}
	\centering
	\includegraphics[width=\textwidth]{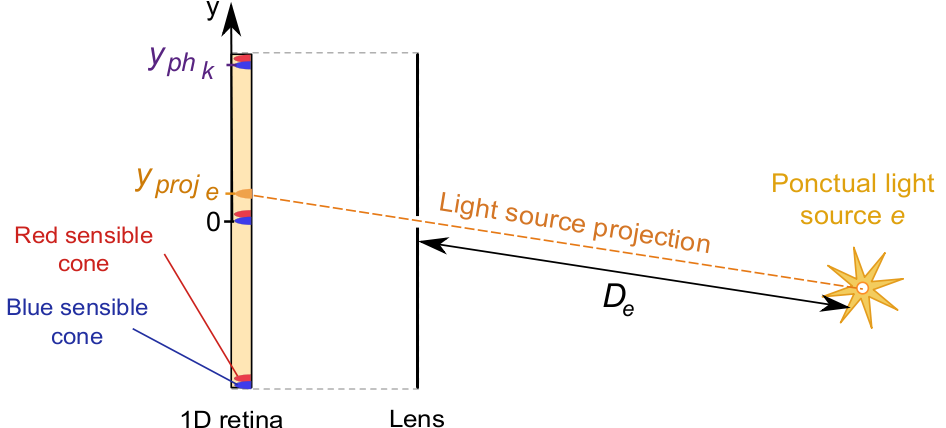}
	\caption{Light projection.}
	\label{fig:projection}
	\end{subfigure}
	\qquad
    \begin{subfigure}[b]{0.43\linewidth}
	\centering
	\includegraphics[width=\textwidth]{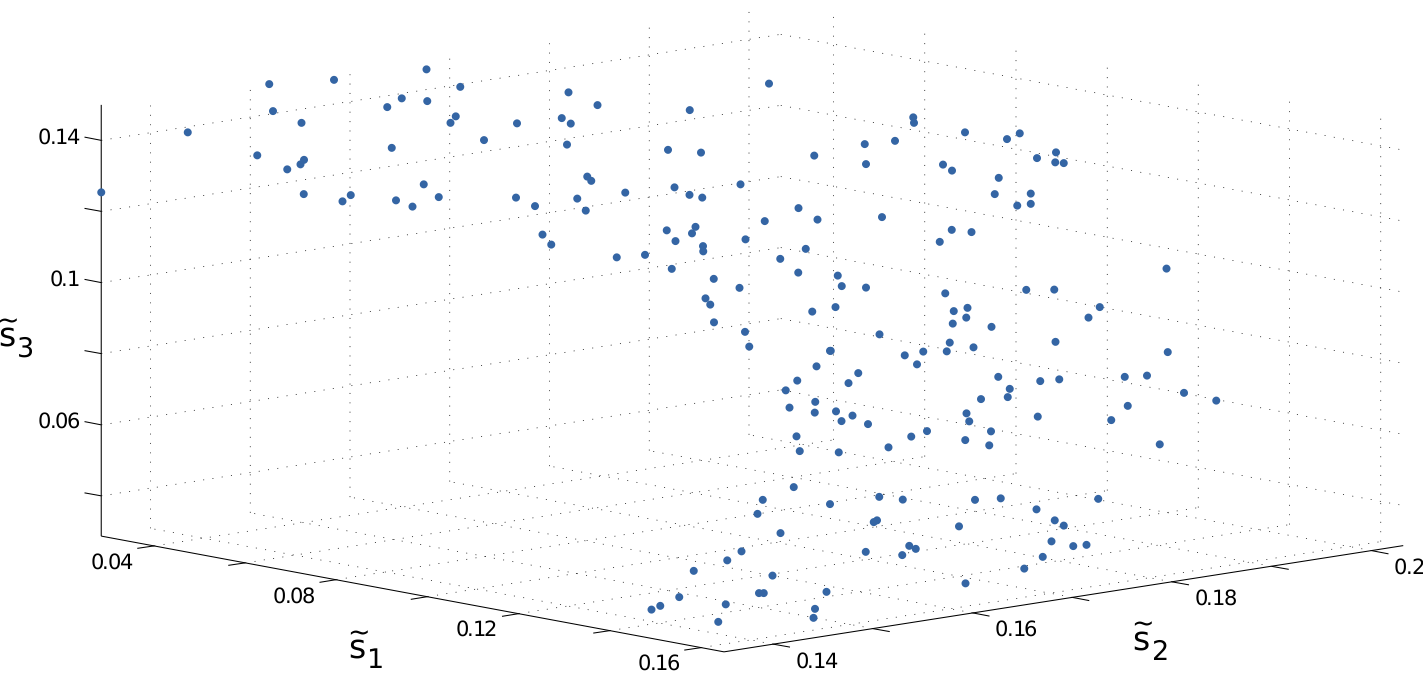}
	\caption{Projection of the exteroceptive dataset.}
	\label{fig:finalsensations}
	\end{subfigure}	
\caption{The robotic arm is made up of three segments of equal length. The relative segments orientations are controlled by individual motors $\{m_1,m_2,m_3\}$. (a) Scenario 1: the end-effector is equipped with a position sensor. (b) Scenario 2: the end-effector is equipped with a retina-like sensor sensible to colored light sources placed in the environment. (c) Each source is projected on the retina through a pinhole lens. The closer the source is to the retina and its projection is to a cone, the stronger the cone excitation. (d) Example of a $6$D exteroceptive dataset projected on the $3$ first coordinates of the exteroceptive space.}
\label{fig:system}
\end{figure*}

\subsection{Jacobian estimation}

Once trained, the NN defines a mapping $F$ from the proprioceptive space $\mathcal{P}$ to the internal representation space $\Xi$, topologically equivalent to the configuration space $\mathcal{C}$ of the end-effector (see Fig.\ref{fig:spaces_and_mappings}).

The mappings $F$ and the corresponding spaces $\Xi$ vary from run to run. What should remain invariant however, is that if any two configurations of the robot $p^1$ and $p^2$ are mapped to the same $\xi$ by the function $F$ computed in the first run, then these configurations must be also mapped to the same value $\xi'$ by the function $F'$ computed in any other trial. Moreover, if two configurations $c^1$ and $c^2$ are close in $\mathcal{C}$, their corresponding representations $\xi^1$ and $\xi^2$ must be close in the internal space $\Xi$.  

This simply implies that the Jacobians of the functions $F$ and $F'$ have the same null-spaces (kernel) at every point $p$ and their null-spaces coincide with the null-space of the mapping
\begin{equation*}
  c = G(p)
\end{equation*} 
relating the configuration of the robot to the real end-effector position.

The Jacobian of $F$ is defined as
\begin{equation}
J_F(p) = \frac{\partial F}{\partial p} = \left(
	\begin{array}{c c c}
		\frac{\partial \xi_1}{\partial p_1} & \dots & \frac{\partial \xi_1}{\partial p_{N^p}} \\
		\vdots & \vdots & \vdots \\
		\frac{\partial \xi_{n}}{\partial p_1} & \dots & \frac{\partial \xi_n}{\partial p_{N^p}}
	\end{array}
	\right).
\label{eq:JF}
\end{equation}

The partial  derivatives $\partial \xi / \partial p_k$ were computed using Euler's formula
\begin{equation}
	\frac{\partial \xi}{\partial p_k} =  \frac{1}{\epsilon}\Bigl(F(p + \epsilon p_k)-F(p)\Bigr)
\label{eq:Jestimation}
\end{equation}
where $p_k$ is the $k$-th basis vector of the proprioceptive space and $\epsilon$ is a small value.

For later comparison, the Jacobian $J_G$ of $G$ is determined analytically.

\subsection{Use of the estimated Jacobian in a reaching task}
\label{sec:reaching}

The trained NN captures the topological properties of the robot's forward kinematics and can then be used to perform a reaching task. Let $p(0)$ be an initial proprioceptive configuration in the portion of the proprioception space $\mathcal{P}$ explored during the learning phase and $\xi(0)$ be the output associated with $p(0)$. Let $\xi^*$ be the desired configuration in the internal representation space $\Xi$. The Jacobian $J_F$ of $F$ can be used to determine the shortest trajectory in the proprioceptive space such that internal state $\xi(t)$ moves along a straight line from $\xi(0)$ to $\xi^*$. The reaching algorithm is as follows, with $t$ the current iteration:
\begin{itemize}
	\item estimate the Jacobian in $p(t)$ (see (\ref{eq:JF}), (\ref{eq:Jestimation})).
	\item compute the proprioceptive configuration update:
		\begin{equation}
		\label{eq:motorupdate}
		\Delta p(t) = J_F^+ \Delta \xi = J_F^+ (\xi^*-\xi(t)),
		\end{equation}
		where $J_F^+$ denotes the Moore-Penrose pseudoinverse of $J_F$.
	\item update proprioceptive configuration with a fixed step size $\epsilon$:
		\begin{equation}
		p(t+1) = p(t) + \epsilon \frac{\Delta p(t)}{||\Delta p(t)||}
		\end{equation}
	\item iterate the process with $t = t+1$, while $||\xi(t) - \xi^*||>\epsilon$.
\end{itemize}

Similar rules were used when reaching was performed in the real end-effector configuration space $\mathcal{C}$ using the analytic Jacobian $J_G$. We call the reaching trajectories `$F$-based' when they are generated using the NN-estimated function $F$, and `$G$-based' if the exact mapping $G$ was used instead.

\section{RESULTS}

The learning and reaching algorithms described in \ref{sec:Neural network and cost function} and \ref{sec:reaching} are applied to a simulated robotic arm in 2D space (see Fig. \ref{fig:arm}). The robot is redundant and has access to three proprioceptive variables $p_k$ when the end-effector has only two degrees of freedom (dof).
Two scenarios are considered depending on the kind of exteroceptive information the robot has access to (described hereafter).
For each scenario, the algorithms parameters are set to:
\begin{itemize}
\item threshold of the neighborhood function: $\mu = 0.1$.
\item centering factor: $\gamma=10^{-3}$.
\item maximum number of learning iterations: $T=1500$.
\item minimal error value: $Q_{min} = 10^{-10}$.
\item number of exploratory movements: $N=1000$.
\item reference proprioceptive configuration:\\$[m_1,m_2,m_3]=[\frac{\pi}{5},-\frac{3\pi}{5},\frac{3\pi}{5}]$ (rad).
\item proprioceptive exploratory amplitude: $A=\frac{\pi}{2}$ (rad).
\item proprioceptive amplitude for the Jacobian estimation and update step size: $\epsilon=10^{-3}$.
\end{itemize}

\subsection{Scenario 1}

In the first scenario, the robot is equipped with a position sensor providing the cartesian coordinates of its end-effector in absolute space. 
The one redundant dof of the arm implies that the null-spaces of $J_F$ and $J_G$ are lines in proprioceptive space $\mathcal{P}$ from which angular divergence can be computed.
On a set of 49 proprioceptive configurations $p^i$ associated with regularly distributed end-effector positions $c^i$ in the working space (see Fig. \ref{fig:bigfig}a and \ref{fig:bigfig}d), those null-spaces were close with an angular divergence of $1.5\deg\pm 1.5\deg$ (average $\pm${} standard deviation).


Figure \ref{fig:bigfig}a presents several $F$-based and $G$-based reaching trajectories computed using the mappings $F$ and $G$ respectively. The target points for $F$- and $G$-based trajectories are defined in different spaces ($\Xi$ and $\mathcal C$, respectively). To assure that they are consistent with each other, we first choose the target point $c^*$ for the $G$-based trajectory, take a random robot configuration $p^*$ corresponding to this point (i.e. $G(p) = c$), and then define the target $\xi^*$ as $\xi^* = F(p^*)$ for $F$-based trajectory.

Note that the configuration $p^*$ usually does not coincide with the final configuration of the arm for the $F$-based trajectory. Hence, if the mapping $F$ were imprecise, the final points of the $F$- and $G$-based trajectory could be different. As Fig.~\ref{fig:bigfig}a clearly shows, it is not the case here.

There is still a small divergence between the $F$- and $G$-based trajectories. The reason of it is that the spaces $\Xi$ and $\mathcal C$ do not have the same metrics and hence a straight line in $\mathcal C$ may correspond to a curve in $\Xi$ and \textit{vice versa}. Figure~\ref{fig:bigfig}b shows the same reaching trajectories, but in the space of internal representation $\Xi$. It can clearly be seen that the $F$-based trajectories, which were curved in the space $\mathcal C$ (Fig~\ref{fig:bigfig}a), are straight in the space of $\Xi$, and that the opposite holds for $G$-based trajectories.

The proprioceptive (angular) profiles differed very slightly for $F$- and $G$-based trajectories. Figure~\ref{fig:bigfig}c presents the worst case (corner-to-corner reaching) and still the blue and orange curves are barely distinguishable. Taken together, all presented evidences suggest that in scenario 1 the robot was able to learn the internal representation of its end-effector configuration space.

\subsection{Scenario 2}

In the second scenario, the robot is equipped with an array of photoreceptors resembling animal's retina. To keep the dimensionality of the problem the same as in the first scenario we assume that the retina preserves its orientation in the absolute reference frame. Note that this assumption is made exclusively for the sake of comparison; a qualitatively similar behavior can be expected when the retina is allowed to rotate.

\begin{figure*}
\centering
\includegraphics[width=.9\linewidth]{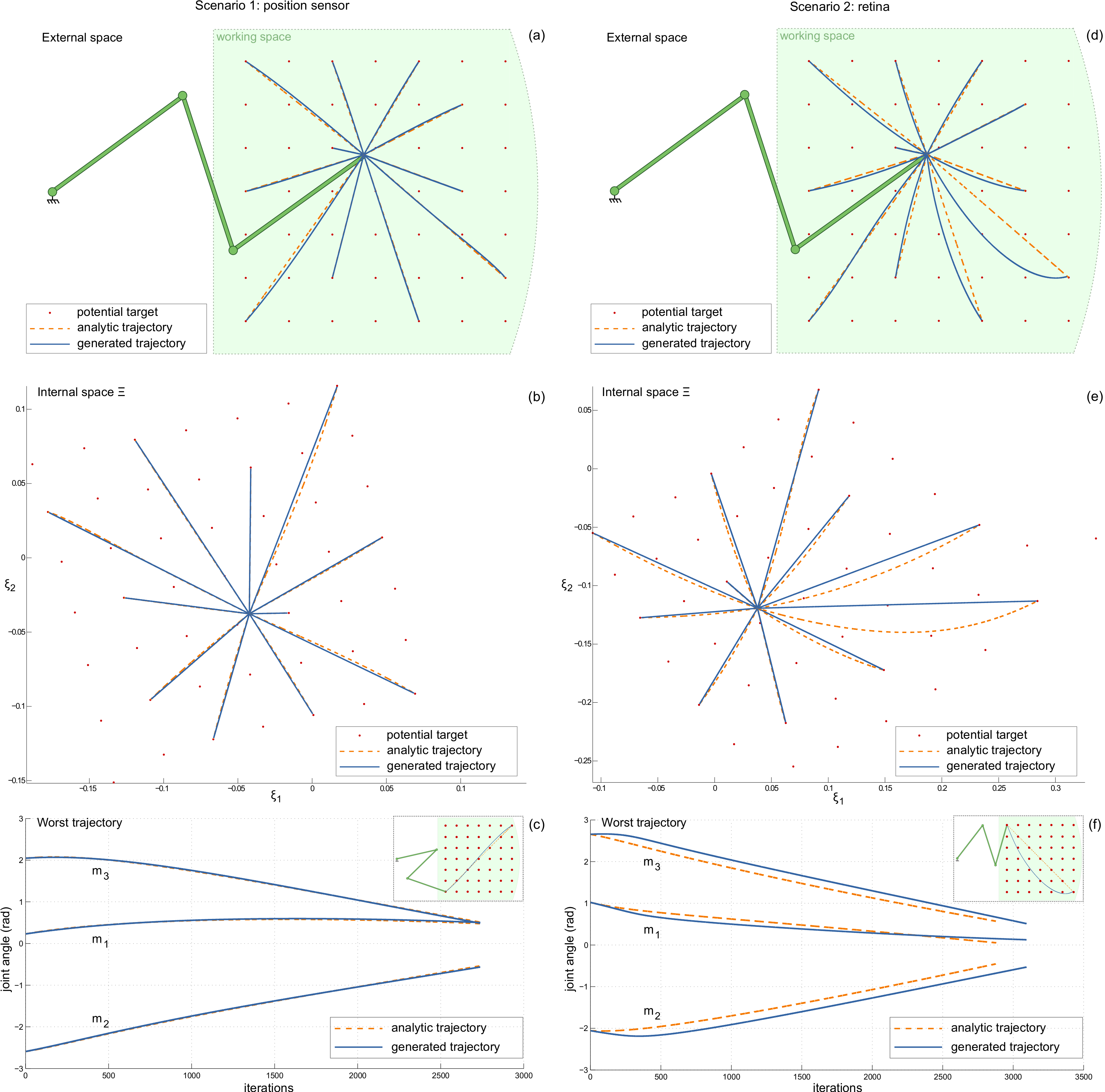}
\caption{(a)(d): $G$-based and $F$-based trajectories for several target configurations $c^*$ distributed in the working space for scenarios 1 and 2 respectively. (b)(e): internal target states $\xi^*$ associated with external configurations $c^*$ and $G$-based and $F$-based trajectories in the internal space $\Xi$. (c)(f): proprioceptive profiles of the worst reaching case task displaying the greatest deviation of $F$-based trajectory from the $G$-based one.}
\label{fig:bigfig}
\end{figure*}
The environment is represented by numerous red and blue light sources placed in front of the robot (see Fig. \ref{fig:arm2}). The light sources are projected on the retina through a pinhole lens as illustrated in Fig. \ref{fig:projection}. There are two kinds of photoreceptors: sensitive to blue light and sensitive to red light. The retina has six photoreceptors, three of each kind. The response of the $k$-th photoreceptor to a single source $e$ has Gaussian profile:
\begin{equation}
s_{k,e} = \frac{\text{exp}(-|| y_{proj_e} - y_{ph_k} ||^2)}{D_e}
\end{equation}
where $y_{proj_e}$ is the position of the source $e$ projection on the retina, $y_{ph_k}$ the position of the photoreceptor $k$ on the retina and $D_e$ the distance between the lens' center and the source $e$. The excitation $s_k$ of the $k$-th photoreceptor equals the sum of $s_{k,e}$ for all sources $e$ of the corresponding color.

In this case the exteroceptor space is built on six photoreceptors and thus has dimension six. For illustration, a three-dimensional projection of the exteroception component of the training data  is shown in Fig.~\ref{fig:finalsensations}. In scenario~2, this cloud of points represents the information that the agent has about its end-effector position. One can notice that the points tend to form a surface. This happens because although the dimension of the exteroception space is six, there is only a two-dimensional manifold of different positions of the end-effector (retina). Clearly, from this cloud of points, the position of the end-effector cannot be easily extracted. Moreover, as the environment changes, so does the surface to which the points lean.

Figure~\ref{fig:bigfig}d presents the $F$- and $G$-based trajectories generated for the same reaching targets, defined the same way as in scenario~1. Again, in spite of the fact that the angular configurations are different at the final points of both types of trajectories (see Fig.~\ref{fig:bigfig}f), the final end-effector positions in the real space are the same. This shows that the mapping $F$ correctly associates the configurations with the same retina position. Indeed, the null-space of the Jacobian $J_F$ was close to the null-space of $J_G$: the angular difference between them was $4.8\deg\pm 5.7\deg$.

The $F$- and $G$-based trajectories have significantly different shapes. This result can be understood by looking at the projection of the potential target grid in the internal representation space $\Xi$ (see Fig.~\ref{fig:bigfig}e). The use of the retina implies that the metrics of $\Xi$, locally derived from exteroceptive metrics, is significantly different from the metrics in $\mathcal C$. The grid of targets $\xi^*$ for $F$-based trajectory is non-uniform in the $\Xi$ space.

The evolution of each proprioceptive component (angles) for the worst reaching case is presented in Fig.~\ref{fig:bigfig}f. Here again, these proprioceptive changes correspond to the shortest angular trajectory for which the robot moves along a straight line in the representation space $\Xi$.

\section{CONCLUSION}

In the current study we presented an algorithm that allows a robot to build an internal representation of its end-effector configuration space on the basis of exteroceptive data. The algorithm does not make any specific assumptions regarding the nature of the exteroception, nor it requires any prior knowledge about the robot's geometrical properties. The internal representation captures the topology of the real end-effector configuration space and preserves the invariance of the null-space of the Jacobian between proprioception and real end-effector position.

The internal representation can be used to control the robot and perform reaching tasks. Similar behavior could be achieved by learning directly the relation between the proprioceptive space and the exteroceptive space \cite{Chaumette2008}. However this relation would vary for different environments.
The mapping $F$ could also be learned by projecting exteroceptive data in a low-dimensional representational space \cite{Lee2007} and then estimating a mapping from the proprioceptive space to this projection. Both the projection and the resulting mapping would yet depend on the environment. The NN we proposed overtake those limitations. On one hand, both the exteroceptive data projection and the mapping learning are performed simultaneously through minimization of the cost function. On the other hand, the topology conservation criteria we introduced ensures that the internal representation is independent from the environment but captures the invariant component of successive explorations: the end-effector configuration.

In the further development of this algorithm, we plan to pay attention to the update of the pre-learned mapping following the change of the environment. If the statistics of the environment content is sufficiently large, this may help to reduce the metrics distortion of the internal representation.

The algorithm is to be tested on a real robot. We expect difficulties to arise with the elements of the robot structure interfering with exteroception. As a consequence, certain configurations can be wrongly classified as different.





\bibliographystyle{plain}
\bibliography{iros2013.bib}

\end{document}